\documentclass[journal]{IEEEtran}

\usepackage{float} \restylefloat{figure} 
\usepackage{color} 

\usepackage{graphicx}
\graphicspath{{eps/}}

\usepackage{fixltx2e}
\usepackage{epstopdf}
\usepackage{rotating}
\usepackage{lscape}
\usepackage{bm} 
\usepackage{cite}
\usepackage{url}
\usepackage[small]{caption}
\usepackage{subfig} 
\usepackage{array} 
\usepackage{mathtools}

\usepackage{amsmath}   
\usepackage{amsfonts}
\usepackage{subfloat}
\usepackage{afterpage}
\long\def\symbolfootnote[#1]#2{\begingroup\def\thefootnote{\fnsymbol{footnote}}
	\footnote[#1]{#2}\endgroup}

\begin{document}

\title{{\Large \textbf{An Online Unsupervised Structural Plasticity Algorithm for Spiking Neural Networks}}}

\author{\authorblockN{Subhrajit Roy,~\IEEEmembership{Student Member,~IEEE} and Arindam Basu,~\IEEEmembership{Member,~IEEE}}\\
\thanks{Manuscript received Nov 15, 2014}
\thanks{The authors are with the School of Electrical and Electronic Engineering,
Nanyang Technological University, Singapore 639798.(e-mail:arindam.basu@ntu.edu.sg). This work was supported by MOE through grants RG 21/10 and ARC 8/13.}
\thanks{Copyright (c) 2010 IEEE. Personal use of this material is
permitted. However, permission to use this material for any other
purposes must be obtained from the IEEE by sending an email to
pubs-permissions@ieee.org.} }
\maketitle
\begin{abstract}
In this article, we propose a novel Winner-Take-All (WTA) architecture employing neurons with nonlinear dendrites and an online unsupervised structural plasticity rule for training it. Further, to aid hardware implementations, our network employs only binary synapses. The proposed learning rule is inspired by spike time dependent plasticity (STDP) but differs for each dendrite based on its activation level. It trains the WTA network through formation and elimination of connections between inputs and synapses. To demonstrate the performance of the proposed network and learning rule, we employ it to solve two, four and six class classification of random Poisson spike time inputs. The results indicate that by proper tuning of the inhibitory time constant of the WTA, a trade-off between specificity and sensitivity of the network can be achieved. We use the inhibitory time constant to set the number of subpatterns per pattern we want to detect. We show that while the percentage of successful trials are 92\%, 88\% and 82\% for two, four and six class classification when no pattern subdivisions are made, it increases to 100\% when each pattern is subdivided into 5 or 10 subpatterns. However, the former scenario of no pattern subdivision is more jitter resilient than the later ones.
\end{abstract}


\IEEEpeerreviewmaketitle
\section{Introduction and Motivation}
The WTA is a computational framework in which a group of recurrent neurons cooperate and compete with each other for activation. The computational power of WTA \cite{Riesenhuber99,NIPS1999_1636,journals/neco/Maass00} and its function in cortical processing \cite{Riesenhuber99,Itti1998} have been studied in detail. Various models and hardware implementations of WTA have been proposed for both rate \cite{Kaski1994,Barnden1993,He1993,Starzyk1993,Serrano1995,Indiveri1997,IndiveriWTA2001,Liu2002} and spike based \cite{Oster2009,McKinstry2013,Masquelier2009} neural networks. In recent past, researchers have looked into the application of STDP learning rule on WTA circuits. The performance of competitive spiking neurons trained with STDP has been studied for different types of input such as discrete spike volleys \cite{Delorme2001,Guyonneau2004487,Masque2007}, periodic inputs \cite{gerstner1993,Yosh2001} and inputs with random intervals \cite{Masquelier2009,Nessler2013,Kappel2014}. 

In this paper, for the first time we are proposing a Winner-Take-All (WTA) network which uses neurons with nonlinear dendrites (NNLD) and binary synapses as the basic computational units. This architecture, which we refer to as  Winner-Take-All employing Neurons with NonLinear Dendrites (WTA-NNLD), uses a novel branch-specific Spike Timing Dependent Plasticity based Network Rewiring (STDP-NRW) learning rule for its training. We have earlier presented \cite{Hussain2014_nc} a branch-specific STDP rule for batch learning of a supervised classifier constructed of NNLDs. The primary differences of our current approach with \cite{Hussain2014_nc} are:
\begin{itemize}
	\item We present an unsupervised learning rule for training a WTA network.
	\item We propose an online learning scheme where connection modifications occur after presentation of each pattern.
\end{itemize}
In this article we consider spike train inputs with patterns occurring at random order which is the same type presented in \cite{Masquelier2009}. The primary differences between our work and the one proposed in \cite{Masquelier2009} are:
\begin{itemize}
	\item Our WTA network is composed of neurons with nonlinear dendrites instead of traditional neurons with no dendrites.
	\item Unlike the network proposed in \cite{Masquelier2009} that requires high resolution weights, the proposed network uses low-resolution non-negative integer weights and trains itself through modifying connections of inputs to dendrites. Hence, change of the `morphology' or structure of the neuron (in terms of connectivity pattern) reflects the learning. This results in easier hardware implementation since a low-resolution non-negative integer weight of $W$ can be implemented by activating a shared binary synapse $W$ times through time multiplexing schemes like address event representation (AER)\cite{Boahen2000,Brink2013}.
	\item In \cite{Masquelier2009}, though the neurons were allowed to learn and respond to subpatterns, there was no actual guideline or control parameter to set the number of subpatterns to	be learned. Here we utilize the slow time constant of the inhibitory signal to select the number of subpatterns we want to divide a pattern into.
\end{itemize}

In the following section, we will present a overview of NNLD, propose the WTA-NNLD architecture and STDP-NRW learning rule and show how the inhibitory slow time constant can be used to select subpatterns within a pattern. Then we shall provide guidelines on selecting the parameters associated with WTA-NNLD and STDP-NRW. In Section IV we will describe the classification task considered in this article which will be followed by the results. We will also present the robustness of the proposed method to variations of parameters in Section V, a quality that is essential for its implementation in low-power, subthreshold neuromorphic designs that are plagued with mismatch. We will conclude the paper by discussing the implications of our work and future directions in the last section.    

\section{Background and Theory}
In this section, we shall first present the working principle of a NNLD. This will be followed by a description of the WTA-NNLD architecture and STDP-NRW learning rule. Lastly, we will throw some light on the role of inhibitory time constant in balancing the specificity and sensitivity of the network. 
\subsection{Neuron with nonlinear dendrites (NNLD)}
The computational model of NNLD was first proposed by Mel et al. in \cite{Mel2001}. They showed that such neurons have higher storage capacity than their non-dendritic counterparts. They used two such NNLDs to construct a supervised classifier and demonstrated its performance in pattern memorization. Recently NNLD has also been employed to develop computationally powerful rate \cite{shaista_ijcnn1} and spike based \cite{roy_biocas1,roy_tbcas} supervised classifiers. The structure of NNLD is isomorphic to a feedforward spiking neural network with a single layer of hidden neurons and one output neuron \cite{Jadi2014}. The lumped dendritic nonlinearities $b()$ are equivalent to the hidden neurons interposed between the input and output layers. However, spiking neurons implement nonlinear thresholding, integration, refractory period etc. Hence, it is typically a much larger circuit compared to the square law nonlinearity of a dendrite, which makes the NNLD an area efficient architecture.
\begin{figure}
	\begin{center}
		\includegraphics[width=0.5\textwidth]{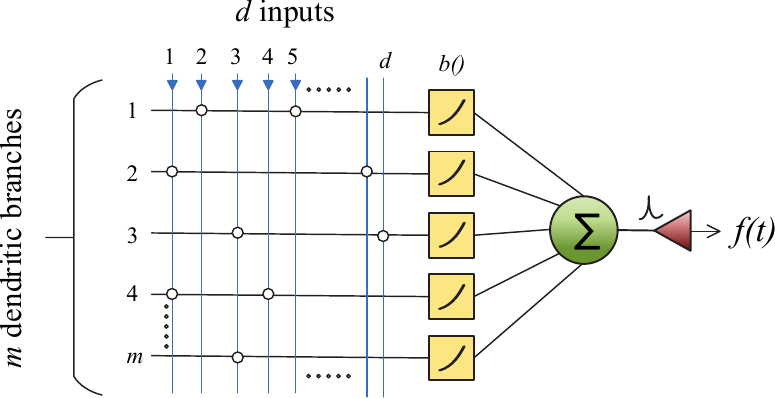}
		\caption{A neuronal cell with nonlinear dendrites\label{fig:Model_NL}}\label{fig:NNLD}
	\end{center}
\end{figure}

As depicted in Fig. \ref{fig:NNLD}, a NNLD consists of $m$ dendritic branches having lumped nonlinearities, with each branch containing $k$ excitatory synaptic contact points of weight 1. If we consider a $d$ dimensional input pattern, then each synapse is driven by any one of these input dimensions where $d>>k$. We use the Leaky Integrate-and-Fire (LIF) model to generate output spikes. Thus, the neuronal membrane voltage is guided by the following differential equation:
\begin{equation}
\label{eq:LIF}
\begin{aligned}
& C\frac{dV(t)}{dt}+\frac{V(t)}{R}=I_{in}(t) \\
& \textrm{If} \;\; V(t)\geq V_{thr}, V(t) \rightarrow 0; \& f(t) \rightarrow 1 \\
& \textrm{else} \; \; f(t)=0
\end{aligned}
\end{equation}
where $V(t)$, $V_{thr}$, $I_{in}(t)$ and $f(t)$ are the membrane voltage, threshold voltage, input current and output spikes of the NNLD respectively. Let us denote the input spike train arriving at the $i^{th}$ input line as $e^{i}(t)$ which is given by: 

\begin{equation}
\label{eq:sp_synap}
e^{i}(t)= \sum\limits_{g} \delta (t-t_{g}^{i})
\end{equation} 
where $g=1,2,...$ is the label of the spike. Then, the input current $I_{in}(t)$ to the neuron can be calculated as:
\begin{eqnarray}
\label{eq:neu}
& I_{in}(t)=\sum\limits_{j=1}^m I_{b,out}^j(t) \\
& I_{b,out}^j(t) = b(I_{b,in}^j(t)) \\
& I_{b,in}^j(t)= \sum\limits_{i=1}^{d} w_{ij}(\sum\limits_{t_{g}^{i} < t } K(t-t_{g}^{i}))
\end{eqnarray} 
where $b(.)$ is the nonlinear activation function of the dendritic branch characterized by $b(z)=z^2/x_{thr}$, $I_{b,out}^j(t)$ is the output current of the $j^{th}$ dendrite and $I_{b,in}^j(t)$ is the input current to the dendritic nonlinearity. Here, $x_{thr}$ describes the behavior of the dendritic nonlinear function by setting a limit on the minimum number of coactive synapses required to produce a supra-linear response. $K$ denotes the post-synaptic current kernel given by:
\begin{equation}
K(t)=I_0(e^{-\frac{t}{\tau_s}}-e^{-\frac{t}{\tau_f}})
\end{equation}
where $\tau_f$ and $\tau_s$ are the fast and slow time constants governing the rise and fall  times respectively and $I_0$ is a normalizing constant. In this article, we consider low resolution non-negative integer weights associated with the input lines. So $w_{ij}\epsilon\{0,1,....,k\}$ means the number of times the $i^{th}$ input line is connected to the $j^{th}$ dendritic branch. Like our earlier work \cite{roy_tbcas}, we allow multiple connections of one input dimension to a single dendrite but restrict the number of connections per dendrite to $k$ by enforcing $\sum_{i=1}^{d} w_{ij} = k$ for each $j$. The output of the NNLD is a spike train and can be denoted as 
\begin{equation}
\label{eq:output_spike}
f(t)= \sum\limits_{a} \delta (t-t_{a})
\end{equation}
where $a=1,2,...$ is the spike index.

\subsection{Winner-Take-All employing Neurons with NonLinear Dendrites (WTA-NNLD)}
\begin{figure}
	\begin{center}
		\includegraphics[width=0.5\textwidth]{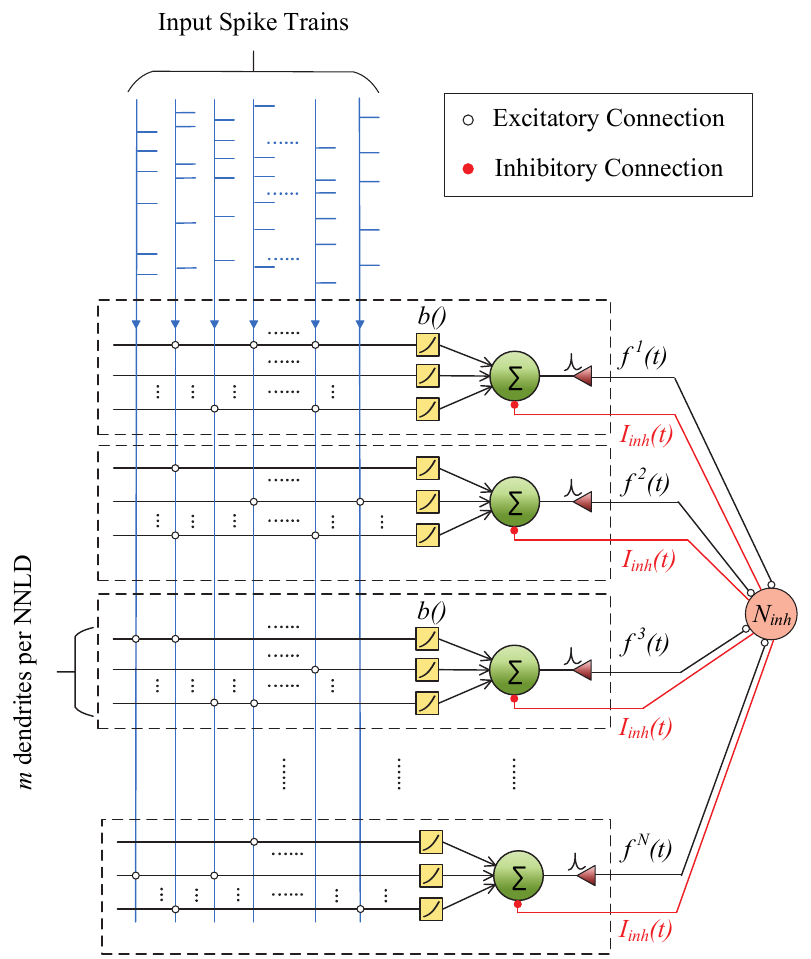}
		\caption{A spike based WTA network employing neurons with lumped dendritic nonlinearities as the competing entities. For implementing lateral inhibition, a inhibitory neuron has been included which, upon activation, provides a global inhibition signal to all the NNLDs.}
		\label{fig:wta_model}
	\end{center}
\end{figure}

We propose a spike based WTA network, depicted in Fig. \ref{fig:wta_model}, which is composed of $N$ such NNLDs. Each NNLD is composed of $m$ dendrites, where each dendrite chooses (repetition allowed) $k$ of the $d$ available input lines and connects to $k$ synapses having weight 1. The membrane voltage, threshold voltage and input current of the $n^{th}$ NNLD are denoted by $V^n(t)$, $V_{thr}$ and $I^n_{in}(t)$ respectively and their dynamics is governed by Equation \ref{eq:LIF}. For the $n^{th}$ NNLD, while the pre-synaptic spike-train arriving at the $i^{th}$ input line is denoted by $e^{i}(t)$ as before, the emitted output spike train is given by $f^n(t)=\sum\limits_{a} \delta (t-t_a^{n})$. Note that for any applied input pattern, $t_a^{n}$ is measured from its beginning.   

We have modelled the effect of lateral inhibition by providing each NNLD with a global inhibitory current signal $I_{inh}(t)$ supplied by a single inhibitory neuron $N_{inh}$ through synapses. The signal $I_{inh}(t)$ is provided by the inhibitory neuron to all the NNLD whenever any one of them fires an output spike. $I_{inh}(t)$ is modeled as $I_{inh}(t)=K_{inh}(t-t_{last}^{n})$, when the last post-synaptic spike is produced by the $n^{th}$ NNLD at $t_{last}^{n}$. The inhibitory post-synaptic kernel, $K_{inh}$, is  given by: 
\begin{equation}
K_{inh}(t)=I_{0,inh}(e^{-\frac{t}{\tau_{s,inh}}}-e^{-\frac{t}{\tau_{f,inh}}})
\end{equation}
where $\tau_{f,inh}$ and $\tau_{s,inh}$ are the fast and slow time constants dictating the rise and fall times of the inhibitory current respectively and $I_{0,inh}$ sets its amplitude. 

\subsection{Spike Timing Dependent Plasticity based Network Re-Wiring learning rule (STDP-NRW)}
\label{sec:STDP-NRW}
\begin{figure}
	\begin{center}
		\includegraphics[width=0.5\textwidth]{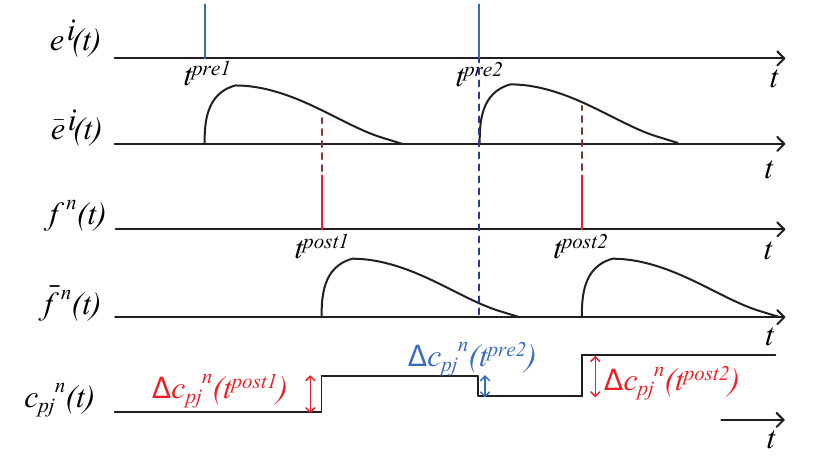}
		\caption{An example of the update rule of fitness value ($c_{pj}^{n}(t)$) is shown. When a post-synaptic spike occurs at $t^{post1}$ the value of $c_{pj}^{n}(t)$ increases by $b_j'(t^{post1})\bar{e}_{i}(t^{post1})$. Due to the appearance of a pre-synaptic spike at $t^{pre2}$, $c_{pj}^{n}(t)$ reduces by $b_j'(t^{pre2})\bar{f}^{n}(t^{pre2})$ as shown in the figure.}
		\label{fig:pre_post_pre}
	\end{center}
\end{figure}

Since we consider binary synapses with weight $0$ or $1$, we do not have the provision to keep real valued weights associated with them. Hence, to guide the unsupervised learning, we define a correlation coefficient based fitness value $c_{pj}^{n}(t)$ for the $p^{th}$ synaptic contact point on the $j^{th}$ dendrite of the $n^{th}$ NNLD of the WTA network, as a substitute for its weight. In the proposed algorithm, structural plasticity or connection modifications happen in longer timescales (at the end of patterns) which is guided by the fitness function $c_{pj}^{n}(t)$ updated by a STDP inspired rule in shorter timescales (at each pre- and post-synaptic spike). The operation of the network and learning process comprises the following steps whenever a pattern is presented:
\begin{itemize}
	\item  $c_{pj}^{n}(t)$ is initialized as $c_{pj}^{n}(t=0)=0 \; \forall \; p=1,2,...k; \;j=1,2,...,m \;\&\; n=1,2,...,N$.
	\item  The value of $c_{pj}^{n}(t)$ is depressed at pre-synaptic and potentiated at post-synaptic spikes according to the following rule:
	 \begin{enumerate}
	 	\item Depression: If the pre-synaptic spike occurs at the $p^{th}$ synapse on the $j^{th}$ dendritic branch of the $n^{th}$ NNLD at time $t^{pre}$, then the value of $c_{pj}^{n}(t)$ at $t=t^{pre}$ is updated by a quantity $\Delta c_{pj}^{n}(t=t^{pre})$ given by:
	 	\begin{equation}
	 	\left. \Delta c_{pj}^{n}(t)=-b_j'(t)\;\bar{f}^{n}(t)\; \right\vert_{t=t^{pre}}
	 	\end{equation}
	 	where $\bar{f}^{n}(t)= K(t) \ast f^n(t)$ is the post-synaptic trace of the $n^{th}$ NNLD and $b'()$ denotes derivative of the nonlinear function $b()$.
	 	\item Potentiation: If the $n^{th}$ NNLD of the WTA-NNLD network fires a post-synaptic spike at time $t^{post}$ then $c_{pj}^{n}(t)$ at $t=t^{post}$ $\; \forall \; p=1,2,...k; \;j=1,2,...,m$  i.e. for each synapse connected to the $n^{th}$ NNLD is updated by $\Delta c_{pj}^{n}(t=t^{post})$ given by:
	 	\begin{equation}
	 	\left. \Delta c_{pj}^{n}(t)=b_j'(t)\;\bar{e}^{i}(t) \; \right\vert_{t=t^{post}} 
	 	\end{equation}
	 	where $\bar{e}^{i}(t) = K(t) \ast e^{i}(t)$ is the pre-synaptic trace of the corresponding input line connected to it.
	 \end{enumerate}
	 A pictorial explanation of this update rule of $c_{pj}^{n}(t)$ is shown in Fig. \ref{fig:pre_post_pre}. Note that for a square law nonlinearity, $b'(z) \propto z$ and hence can be easily computed in hardware without requiring any extra circuitry to calculate the derivative.
	 \item During the presentation of the pattern whenever a spike is produced by any of the $N$ excitatory NNLDs, the inhibitory neuron $N_{inh}$ sends an inhibitory signal to all the NNLDs of the WTA.
	 \item After the network has been integrated over the current pattern of duration $T_p$, the synaptic connections of the NNLDs which have produced at least one spike are modified. 
	 \item If we consider that $Q$ out of $N$ NNLDs have produced post-synaptic spike/spikes for the current pattern, then the connectivity of the $q^{th}$ NNLD $\forall q=1,2...,Q$ is updated by tagging the synapse ($s_{min}^{q}$) having the lowest value of correlation coefficient at $t=T_p$ out of the $m\times k$ synapses connected to it for possible replacement.    
	 \item To aid the unsupervised learning process, randomly chosen sets $R^{q}$ containing $n_R$ of the $d$ input dimensions are forced to make silent synapses of weight 1 on the dendritic branch of $s_{min}^{q}$ $\forall \; q=1,2...,Q$. We term these synapses as ``silent" since they do not contribute to the computation of $V^n(t)$ - so they do not alter the classification when the same pattern set is re-applied. The value of $c_{pj}^{q}(t=T_p)$ is calculated for synapses in $R^{q}$ and the synapse having maximum $c _{pj}^{q}(t=T_p)$ in $R^{q}$ denoted by $r_{max}^{q}$  $\forall \; q=1,2...,Q$ is identified. Next, the input line connected to $s_{min}^{q}$ is swapped with the input line connected to $r_{max}^{q}$. Hence, instead of the traditional method of training by changing of high-resolution synaptic weights, our learning rule modifies the connections between the inputs and dendrites based on the fitness values.  
	 \item All the $c^n_{pj}(t)$ values are reset to zero and the above mentioned steps are repeated whenever a pattern is presented. Here, we define an epoch for $C$ class classification as a set of patterns consisting of one pattern from each of the $C$ classes- in random order. We define another term $l_{mean}$ as the average of the latencies of the post-synaptic spikes in the network over time period of the last epoch which is given by:
	 \begin{equation}
	 	l_{mean}= <\sum_{n} \sum_{a} t_a^n>_1
	 \end{equation}
	 where $<\cdot>_1$ denotes averaging over one epoch. We note the value of $l_{mean}$ for every epoch and the learning is considered to converge when the value of a `Convergence Measure' ($CM$) based on $l_{mean}$ reaches saturation. We define our `Convergence Measure' in Section \ref{params}. 
\end{itemize}  

\subsection{Specificity and Sensitivity: Role of Inhibitory Time Constant}\label{spec_sens}
When a pattern is presented to the WTA-NNLD and any one of the $N$ NNLDs produce an output spike, a global inhibition current $I_{inh}(t)$ is injected into all the $N$ NNLDs. The slow time constant $\tau_{s,inh}$ of this signal controls the output firing activity of the WTA-NNLD. Typically, a large value of $\tau_{s,inh}$ (w.r.t to $T_p$) is set, and only one NNLD produces an output spike i.e. patterns of same class are encoded by a single NNLD. The post-synaptic spike latency for a pattern $P$ is defined as the time difference between the start of the pattern
and the first spike produced by any one of the $N$ neurons of WTA-NNLD. During training of WTA-NNLD for this case, different NNLDs get locked onto different classes of pattern and
the latency gradually decreases until the end of the training. Thus, after completion of training, the unique NNLDs which have learned different classes of pattern rely only on the first
few spikes (determined by the latency at the end of training) to predict the pattern's class thereby significantly reducing the prediction time \cite{Masquelier2009}. So, the sensitivity of the network
is increased. However, the problems with this approach are:
\begin{itemize}
	\item The percentage of successful classifications can be less due to the strict requirement of different neurons firing based only on first few spikes of different patterns (shown
	in Section \ref{exp_res}).
	\item Though the prediction time of a pattern's class is significantly reduced, this method neglects most part of the pattern after the first few spikes which may lead to a lot
	of false detections.
\end{itemize}

\begin{figure}
	\begin{center}
		\includegraphics[width=0.5\textwidth]{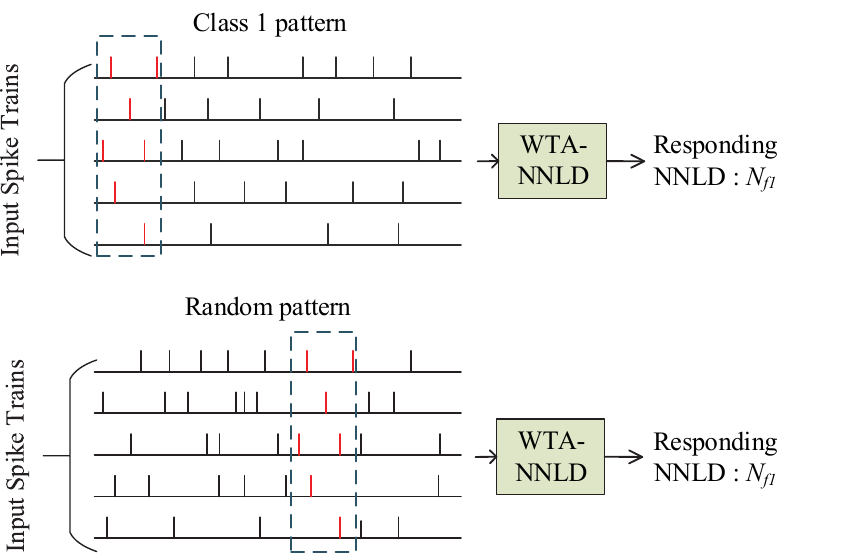}
		\caption{Specificity is reduced if only one NNLD encodes a pattern based on its first few spikes. As shown, a different pattern with a section resembling the beginning of class 1 pattern may cause neuron $N_{f1}$ to respond.}
		\label{fig:eg_nsub1}
	\end{center}
\end{figure}

We demonstrate the limitation mentioned in the above point by a simple example in Fig. \ref{fig:eg_nsub1}. Let us consider we are performing $C$ class classification and assume that after the training phase is complete, NNLD $N_{f1}$ responds to patterns belonging to Class 1. NNLD $N_{f1}$ has trained itself to provide an output spike depending on the position of the first few
spikes (red spikes in dashed box of Fig. \ref{fig:eg_nsub1}) of the pattern. It neglects the rest of the pattern while providing a prediction. However, for longer patterns there is a chance that this spike set can occur anywhere inside a random pattern (not belonging to any class or to another class). The same NNLD $N_{f1}$ responds to such patterns by producing a post-synaptic spike. Thus, we see that though trained WTA-NNLD is very sensitive in this case, it loses specificity. On the other hand, if we set a moderate value of $\tau_{s,inh}$, then for a single pattern multiple NNLDs are capable of producing output spikes. Hence, patterns of the same class are now encoded by a sequence of successive firing of few NNLDs where each NNLD fires for one subpattern. Let $n_{sub}$ be the number of subpatterns that is set by a proper choice of $\tau_{s,inh}$. Thus the original case of one NNLD firing for each pattern corresponds to $n_{sub} = 1$ . In this article, for a $C$ class classification we define a successful trial as one in which (a) during the training phase WTA-NNLD learns different unique representations for patterns of different classes and (b) after completion of training and achieving success in (a), the network produces the same representation, when presented with testing patterns corresponding to classes that it had learned during the training phase. When $n_{sub} = 1$ i.e. no pattern subdivisions are made, this unique representation is a different neuron firing for different classes of patterns. When $n_{sub} >$ 1, the unique representation is a different sequence of successive NNLDs firing for different classes of patterns. When, $n_{sub}>$ 1, we allow the NNLDs to detect subpatterns within patterns. Since in this approach the WTA-NNLD gives weightage to the entire pattern before predicting its class, the number of false detections can be largely reduced. However, this method has a limitation of being less jitter resilient $-$ one of the many subpatterns can be easily corrupt by noisy jitters in spike (shown in Section \ref{exp_res}) and fail to produce a unique identifier during testing phase. Hence, the choice of $n_{sub}$ and consequently the inhibitory time constant depends on the amount of temporal jitter in the application. 

\section{Choice of Parameters}\label{params}

The following is an exhaustive list of the parameters used by WTA-NNLD and STDP-NRW:
\begin{enumerate}
	\item $T_p$: Duration of a pattern
	\item $d$: Dimension of the input
	\item $m$: Number of dendrites per NNLD
	\item $k$: Number of synapses per dendrite
	\item $n_R$: Number of input dimensions in replacement set
	\item $\tau_s$ and $\tau_f$: Slow and fast time constant of excitatory current kernel
	\item $I_0$: Normalization constant of excitatory current kernel
	\item $\tau_{s,inh}$ and $\tau_{f,inh}$: Slow and fast time constant of inhibitory current kernel
	\item $I_{0,inh}$: Normalization constant of inhibitory current kernel
	\item $x_{thr}$: Threshold of dendritic nonlinearity
	\item $V_{thr}$: Firing threshold voltage of NNLD
	\item $N$: Number of NNLDs in WTA
	\item $C$: Number of classes of patterns
\end{enumerate}

We will now provide some guidelines on choosing the key parameters:

\paragraph{Total number of synapses per NNLD (s)}
The number of synapses allocated to each neuronal cell of WTA-NNLD are kept as equal to the dimension ($d$) of the input patterns. This is done to ensure NNLD uses the same amount of synaptic resources as the simplest neuron--a perceptron. Thus, if the proposed network is comprised of $N$ such neuronal cells then the total number of synaptic resources required are $d\times N$.

\begin{figure}
	\begin{center}
		\includegraphics[width=0.4\textwidth]{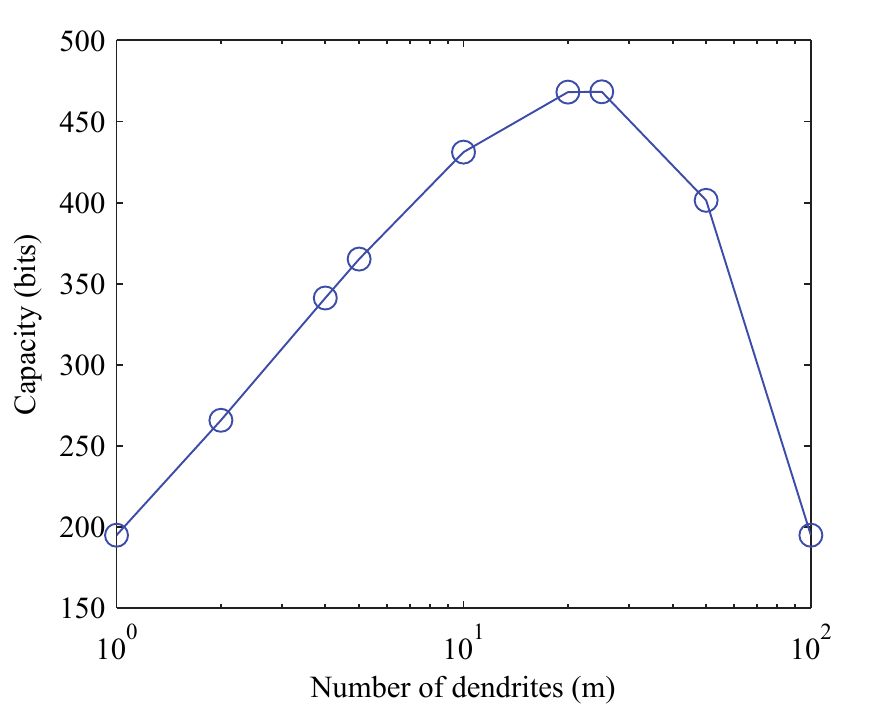}
		\caption{The pattern memorization capacity of a NNLD ($B_N$) is plotted as function of the number of dendrites ($m$) for a fixed number of input dimensions ($d=100$) and synapses ($s=100$).}
		\label{fig:capacity}
	\end{center}
\end{figure}

\paragraph{Number of dendrites per NNLD (m)}
In \cite{Mel2001} a measure of the pattern memorization capacity,$B_{N}$, of the NNLD  (Fig.\ref{fig:Model_NL}) has been provided by counting all possible functions realizable as:
\begin{equation}
\label{eq:BN}
B_{N} = log_2 \binom{\binom{k+d-1}{k}+m-1}{m} bits
\end{equation}
where $m$, $k$ and $d$ are the number of dendrites, number of synapses per dendrites and dimension of the input respectively for this neuronal cell. When a new classification problem is encountered, we first note down the value of $d$, which in turn sets our $s$ since we have considered $s=d$. Since $s=m \times k$, for a fixed $s$ all possible values which $m$ can take are factors of $s$. We calculate $B_N$ for these values of $m$ by Equation \ref{eq:BN}. The value of $m$ for which $B_N$ attains its maxima is set as $m$ in our experiment. As an example, we show in Fig. \ref{fig:capacity} the variation of $B_N$ with $m$ when $d=100$. It is evident from the curve that the capacity is maximum when $m=25$ and so in our simulations for classifying $100$ dimensional patterns we employ neuronal cells having 25 dendrites.

\begin{figure*}[!t]
	\centering
	\subfloat[]{\includegraphics[width=0.32\textwidth]{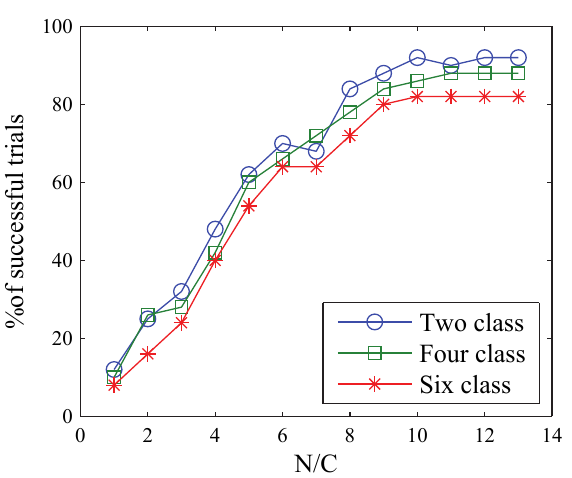}}\
	\subfloat[]{\includegraphics[width=0.32\textwidth]{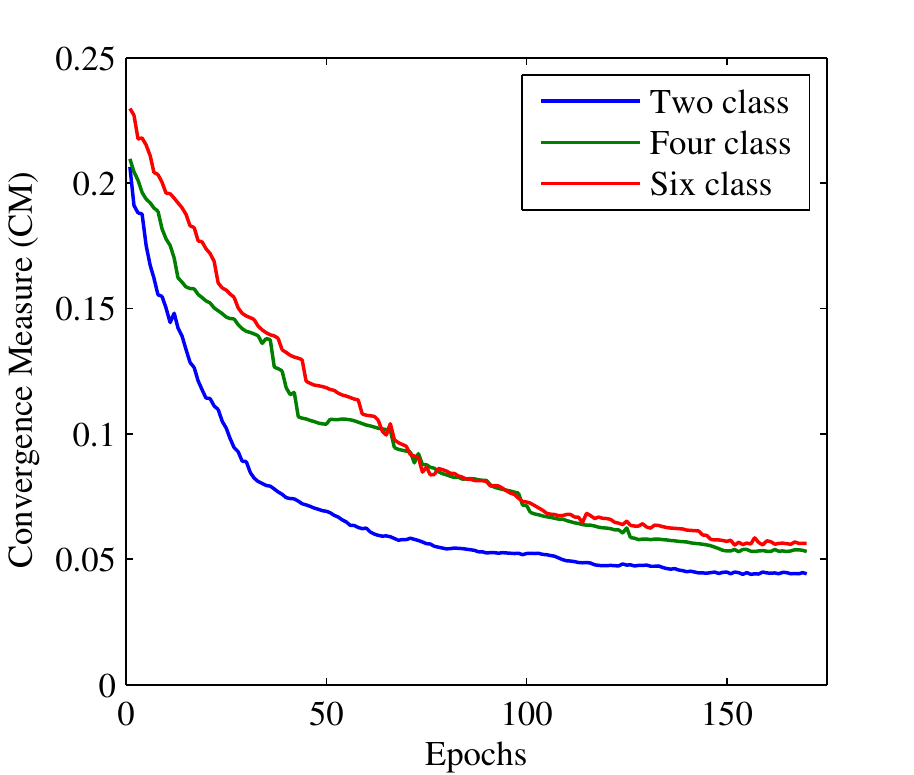}}\
	\subfloat[]{\includegraphics[width=0.32\textwidth]{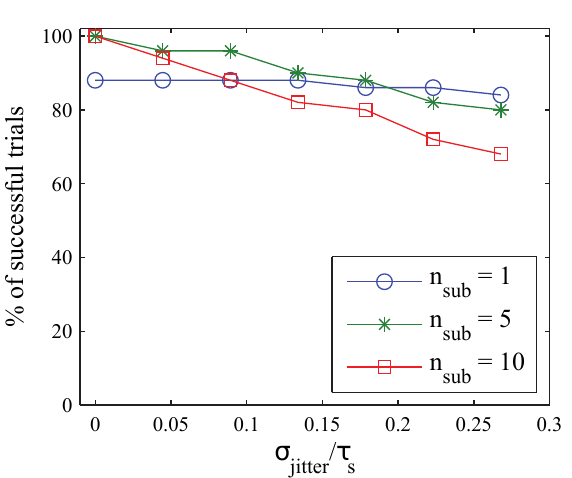}}\
	\caption{(a) The percentage of successful trials is plotted against $\frac{N}{C}$ for two-class, four-class and six-class classification. The figure shows that as $\frac{N}{C}$ increases the percentage of successful trials also increases and becomes constant after $\frac{N}{C}=11$. (b) The evolution of $CM$ (averaged over 50 trials) with the number of epochs for two-class, four-class and six-class classification for $n_{sub}=1$. (c) The percentage of successful trials is plotted against $\sigma_{jitter}/\tau_s$ for $n_{sub}$ =1, 5 and 10. As the number of subpattern divisions are increased the jitter/noise robustness of the network decreases. }
	\label{n_sub_1}
\end{figure*} 

\paragraph{Number of synapses per branch (k)}
After $s$ and $m$ have been set, the value of $k$ can be computed as $k=\frac{s}{m}$.

\paragraph{The normalization constant ($I_0$), slow ($\tau_s$) and fast time constant ($\tau_f$) of excitatory PSC kernel}
The fast time constant ($\tau_f$) and slow time constant ($\tau_s$) have been defined in Section \ref{sec:STDP-NRW}. In hardware implementation of a synapse \cite{roy_tbcas} $\tau_f$ usually takes a small positive value and is typically not tuned. The slow time constant, $\tau_s$, is responsible for integration across temporally correlated spikes and the performance of the network is dependent on its value. If $\tau_s$ takes too small a value, then the post synaptic current due to individual spikes dies down rapidly and thus temporal integration of separated inputs does not take place. On the other hand large values of $\tau_s$ render all spikes effectively simultaneous. So, in both extremes the extraction of temporal features from the input pattern is hampered. In \cite{roy_tbcas} we have provided a mathematical formula for calculating $\tau_{s,opt}$, the optimal value of $\tau_s$, with respect to the inter spike interval (ISI) of the input pattern for which optimal performance of the network is obtained. If we are considering $d$ dimensional patterns and the mean firing rate of each dimension is $\mu_f$, then the mean ISI across the entire pattern is given by $\mu_{ISI}=1/ (d \times \mu_f)$. We can then set $\tau_{s,opt}$ according to the formula:
\begin{equation}
\label{eq:taus_opt}
\tau_{s,opt}=52.83\mu_{ISI}-3.1 \\
\end{equation}

In our simulations, we keep $\tau_f$ as $\tau_f=\frac{\tau_s}{10}$. Since the weights of all the active synapses  are $1$, we set $I_0=1.4351$ to normalize the amplitude of the PSC to be $1$.

\paragraph{Threshold of nonlinearity ($x_{thr}$)}
During the training of WTA-NNLD, the STDP-NRW rule preferably selects those connection topologies where correlated inputs for synaptic connections are connected to the same branch. Thus, the lumped dendritic nonlinearity $b(z)=\frac{z^2}{x_{thr}}$ should give a supra-linear output only when correlated input dimensions are connected to the dendrite. To ensure this we keep the value of $x_{thr}$ equal to the average input to the nonlinear function in case of random connections. We create numerous instances of dendrites having $k$ synapses and calculate the average input to the nonlinear function, $b_{in,avg}$, for the pattern set at hand. Then we set the value of $x_{thr}$ as, $x_{thr}=b_{in,avg}$.        

\paragraph{$V_{thr}$ of NNLD}
The NNLD should provide a post-synaptic spike only when correlated inputs have been connected to its dendrites. We consider a NNLD having $m$ dendrites and $k$ synapses and create numerous instances of random connections to these synapses. We measure the average value of the maximum membrane voltage ($(V_{max})_{av}$) produced when this NNLD is integrated over the pattern duration for all these instances and set $V_{thr} =(V_{max})_{av}$.

\paragraph{The normalization constant ($I_{0,inh}$), slow ($\tau_{s,inh}$) and fast ($\tau_{f,inh}$) time constant of $I_{inh}(t)$}
The post-synaptic firing activity of the WTA-NNLD network is dependent on $\tau_{s,inh}$ and $I_{0,inh}$. To simulate the hardware scenario we set $\tau_{f,inh}$ to a small value given by $\tau_{f,inh}=\frac{\tau_{s,inh}}{10}$. To set $I_{0,inh}$ and $\tau_{s,inh}$, we first excite WTA-NNLD with $ep_{ini}$ epochs of patterns prior to training and calculate the average excitatory current ($I_{e,av}$) to the NNLDs as:
\begin{equation}
	I_{e,av}=< \frac{1}{n} \sum_{n=1}^{N} I^n_{in}(t) >_{ep_{ini}} \\
\end{equation}
where $<\cdot>_{ep_{ini}}$ denotes averaging over $ep_{ini}$ epochs.
The idea is to generate a $I_{inh}(t)$ which, if provided by $N_{inh}$ at the beginning of a subpattern, decays exponentially to $I_{e,av}$ at the end of the subpattern i.e. after time $T_{sub}$ has elapsed. This ensures that once a post-synaptic spike is generated by a NNLD in a particular $T_{sub}$ time window, other NNLDs are unable to fire during that same $T_{sub}$ time window. Assuming $\tau_{f,inh}<<\tau_{s,inh}$, we can derive that the required $I_{inh}(t)$ is implemented by setting $\tau_{s,inh}$ as:
\begin{equation}
\tau_{s,inh}=\frac{T_{sub}}{ln(\frac{I_{0,inh}}{I_{e,av}})} \\
\label{tau_inh_I0_inh}
\end{equation}  

Note that $\tau_{s,inh}$ has an inverse logarithmic relation to $I_{0,inh}$.  

\paragraph{Convergence Measure ($CM$)}
The formula for calculating $CM$, applicable to both $n_{sub}=1$ and $n_{sub}>1$, for detecting the convergence of learning is given by:
	\begin{equation}
	CM = \frac{l_{mean}}{n_{sub}} - \frac{(n_{sub}-1)\; T_{sub}}{2} \\
	\end{equation}

Note that for $n_{sub}=1$, $CM = l_{mean}$	and so it computes the time-to-first spike for patterns averaged over an epoch. For $n_{sub}>1$, $CM$ calculates the average time-to-first spike from the beginning of each subpattern of $C$ patterns of an epoch. We consider the learning has converged when the value of $CM$ saturates. 

\section{Experiments and Results}\label{exp_res}
In this section, we will describe the classification task considered in this article. To show how the classification performance generalizes to multi-class we will consider two, four and six class classification. We will be showing the performance of WTA-NNLD and STDP-NRW for three values of $n_{sub}$ given by $n_{sub}$ = 1, 5 and 10.

\subsection{Problem Description}
The benchmark task we have selected to analyze the performance of the proposed method is the Spike Train Classification problem \cite{Natschlaeger02}. In the generalized Spike Train Classification problem, $C$ arrays of $h$ Poisson spike trains having frequency $f$ and length $T_p$ are present which are labeled as classes 1 to $C$. Jittered versions of these templates are created by altering the position of each spike within the templates by a random amount that is randomly drawn from a Gaussian distribution with zero mean and standard deviation $\sigma_{jitter}$. The network is trained by these jittered versions of spike trains, and the task is to correctly identify a pattern's class. In this article, unless otherwise mentioned, we have considered $h$ = 100 and Poisson spike trains are present in each afferent, $f$ = 20 and $T_p$ = 0.5 sec and varied $C$ and $\sigma_{jitter}$. Inspired by \cite{Masquelier2009}, we also consider the scenario when $h/2$ randomly chosen afferents do not contain any spikes, while the remaining $h/2$ afferents are Poisson spike trains.   

\begin{figure}[!t]
	\centering
	\subfloat[]{\includegraphics[width=0.24\textwidth]{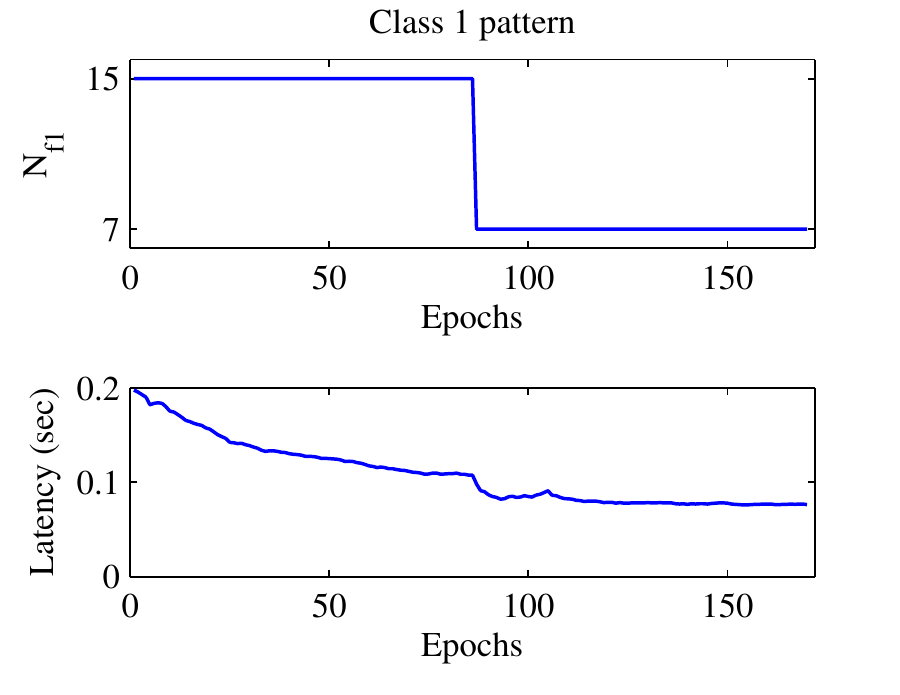}}\
	\subfloat[]{\includegraphics[width=0.24\textwidth]{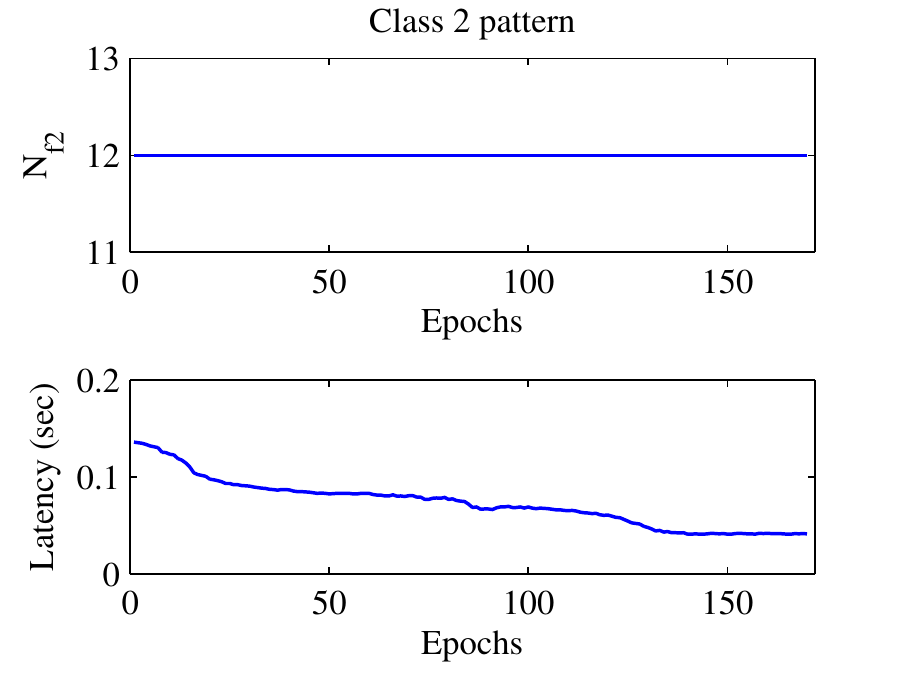}}\\
	\subfloat[]{\includegraphics[width=0.24\textwidth]{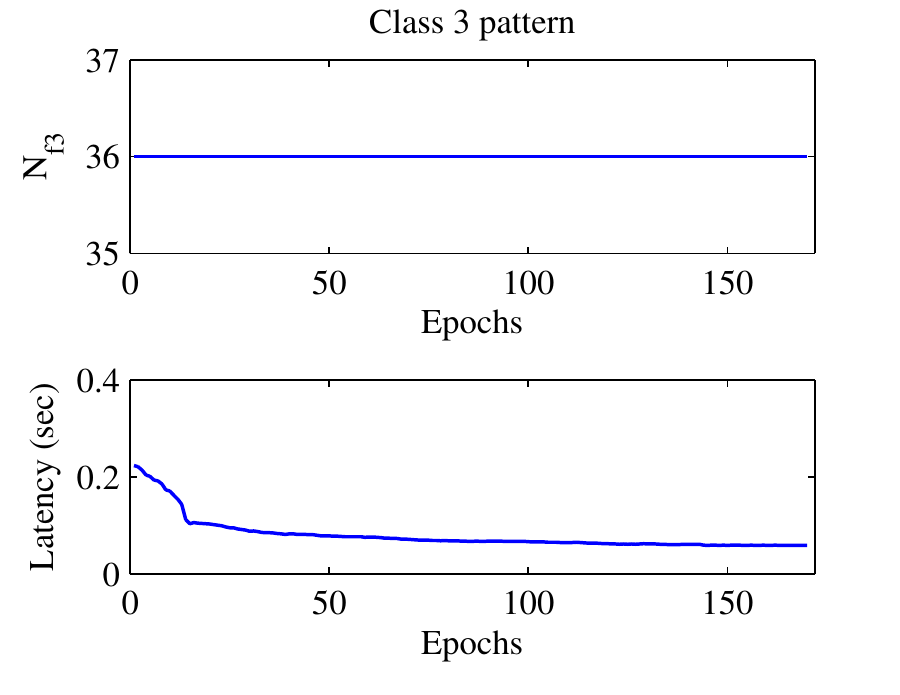}}\
	\subfloat[]{\includegraphics[width=0.24\textwidth]{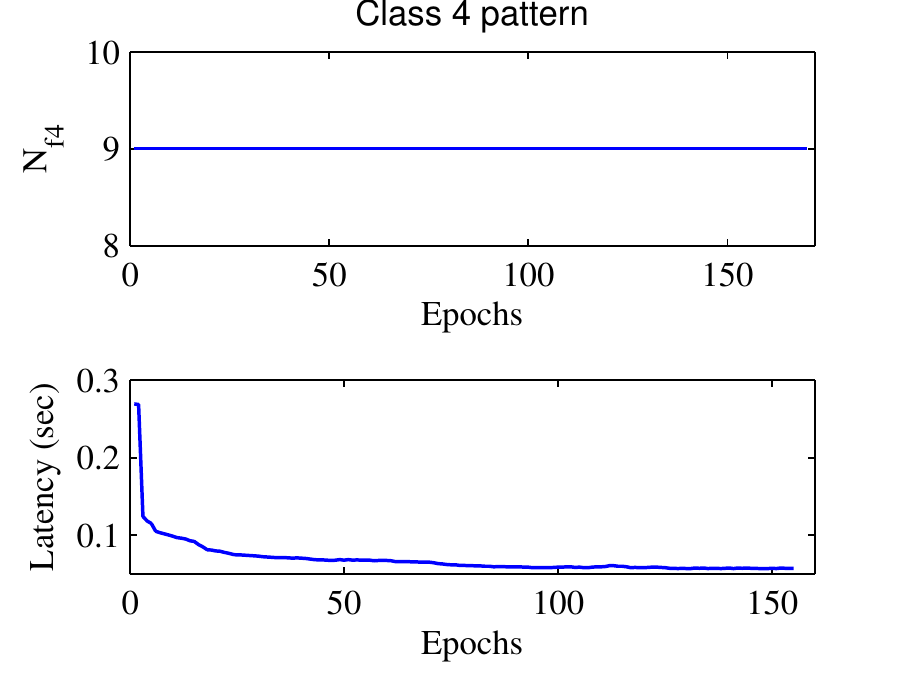}}\
	\caption{For four class classification the above figure shows that out of 44 NNLDs (a) $7^{th}$ NNLD recognizes Class 1 pattern (b) Class 2 pattern fires for the $12^{th}$ NNLD, (c) $36^{th}$ NNLD recognizes Class 3 pattern (d) Class 4 pattern is recognized by $9^{th}$ NNLD and the latency for all four of them decreases over epochs until saturation.}
	\label{fig:nojit_latency_vs_epoch__classes_experi_1}
\end{figure} 

\begin{figure*}[!t]
	\centering
	\subfloat[]{\includegraphics[width=0.32\textwidth]{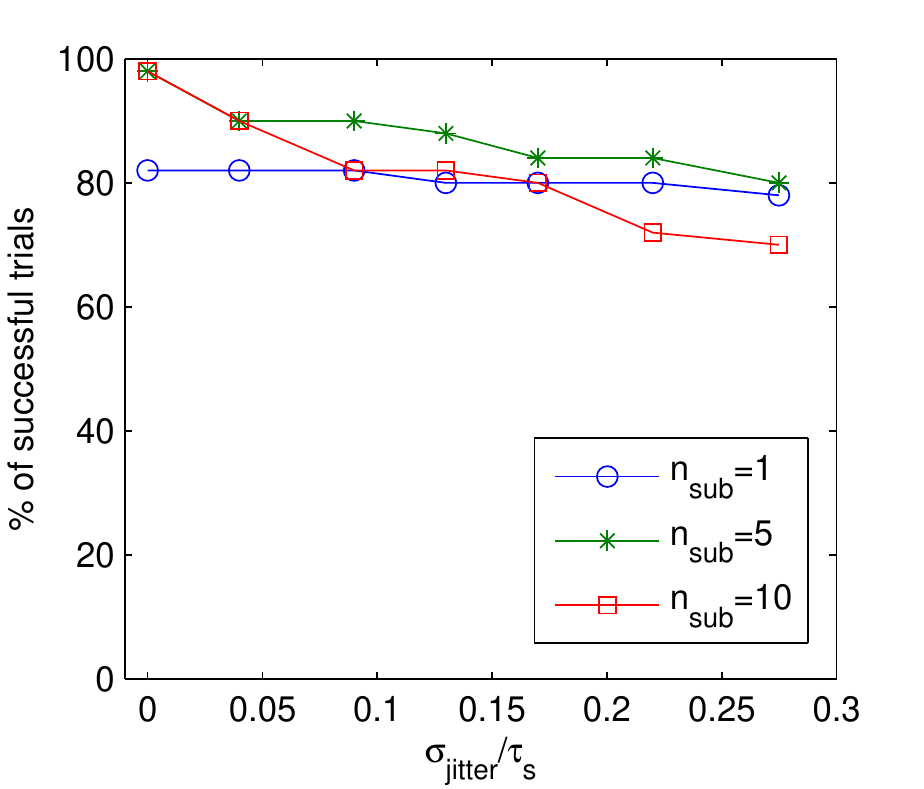}}\
	\subfloat[]{\includegraphics[width=0.32\textwidth]{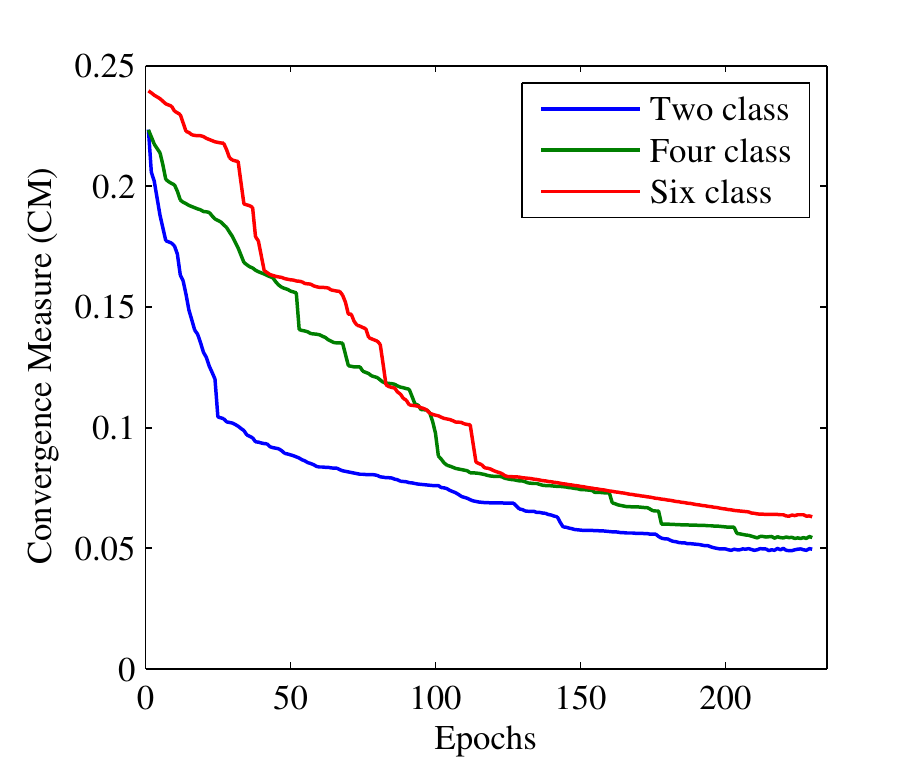}}\
	\subfloat[]{\includegraphics[width=0.32\textwidth]{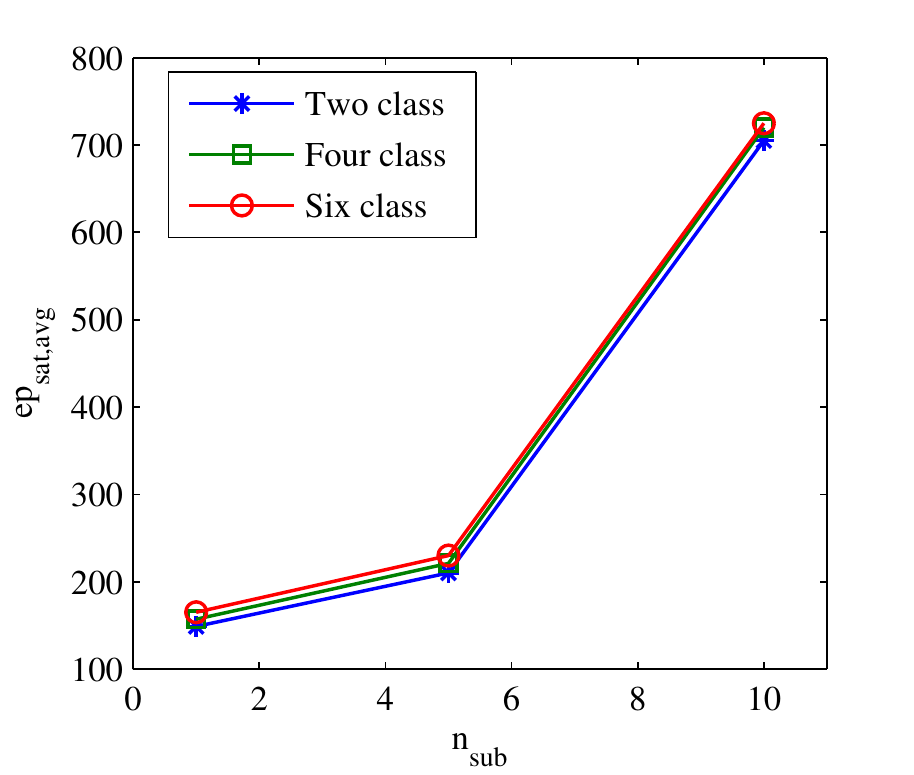}}\
	\caption{(a) The percentage of successful trials is plotted against $\sigma_{jitter}/\tau_s$ for $n_{sub}$ =1, 5 and 10 for patterns having spikes in only 50 \% of the afferents. (b)  The evolution of $CM$ (averaged over 50 trials) with the number of epochs for two-class, four-class and six-class classification when $n_{sub}=5$. (c) This figure depicts the number of epochs needed for saturation of $CM$ (averaged over 50 trials) against the number of subpatterns considered for each pattern ($n_{sub}$). As $n_{sub}$ increases, WTA-NNLD has to train itself for more number of subpatterns and thus there is an increase in $ep_{sat,avg}$. }
	\label{nsub_more_1}
\end{figure*}

\subsection{Case 1: $n_{sub}=1$}
In this case we have $T_{sub} = T_p$, so one NNLD is capable of firing only once when a pattern is presented. Considering $\sigma_{jitter}$ = 0, we have varied the number of NNLDs and noted the percentage of successful trials which is depicted in Fig. \ref{n_sub_1}(a). To make the horizontal axis invariant of the number of classes, we have taken $\frac{N}{C}$ as the horizontal axis. From the figure we can conclude that the percentage of successful trials gradually increases with an increase in $\frac{N}{C}$ and finally becomes constant after $\frac{N}{C} = 11$. Thus, unless otherwise mentioned, we will keep $N = 11 \times C$ when $n_{sub} = 1$. It can be seen from Fig. \ref{n_sub_1}(a) that the percentage of successful trials cannot go beyond 92\%, 88\% and 82\% for two, four and six class classification respectively.

Earlier in Section \ref{params} we have mentioned that learning converges when $CM$ saturates. For $n_{sub}=1$, $CM=l_{mean}$ is the average of the time-to-first spikes for patterns in an epoch. As an example we consider a particular trial of four-class classification and show in Fig. \ref{fig:nojit_latency_vs_epoch__classes_experi_1} that during training, the latencies of the four NNLDs, $N_{f1}$, $N_{f2}$, $N_{f3}$ and $N_{f4}$ which uniquely recognize the four class of patterns gradually reduce until reaching a saturation point. Moreover, in Fig. \ref{n_sub_1}(b) we show the epochwise evolution of $CM$ averaged over 50 trials for two-class, four-class and six-class classification. It is evident from the figure that the value of $CM$ decreases thereby showing that the algorithm is favoring correlated inputs such that the post-synaptic spikes can occur faster and finally saturates after some epochs have passed. We denote the number of epochs taken by the algorithm for saturation of $CM$ as $ep_{sat}$ and note its value for 50 trials. The average value of $ep_{sat}$ for 50 trials, $ep_{sat,avg}$, is then computed to be $ep_{sat,avg}$ = 149, 157 and 165 for two-class, four-class and six-class classification respectively. Moreover, this phenomenon clearly indicates that while the WTA-NNLD network is being trained by STDP-NRW learning rule, $C$ unique NNLDs which have locked onto the $C$ different classes of pattern, are trying to recognize the start of repeating patterns for different classes. Fig. \ref{n_sub_1}(b) also suggests that after the training has stopped, these $C$ unique NNLDs, instead of looking at the whole pattern of duration $T_p = 500 ms$, can now look only at the starting $44.8 ms$, $53.1 ms$ and $56.1 ms$ of the patterns for $C=$ 2, 4 and 6 respectively to predict its class.

Let us now consider the effect of jitter and we show in Fig. \ref{n_sub_1}(c) the performance of the proposed method when the intensity of jitter is varied. Next, we look into the performance of the proposed method when patterns with 50\% empty afferents are considered. Fig. \ref{nsub_more_1}(a) depicts the results obtained by our network for this case with varying amounts of jitter.

\begin{figure*}
	\begin{center}
		\includegraphics[width=0.95\textwidth]{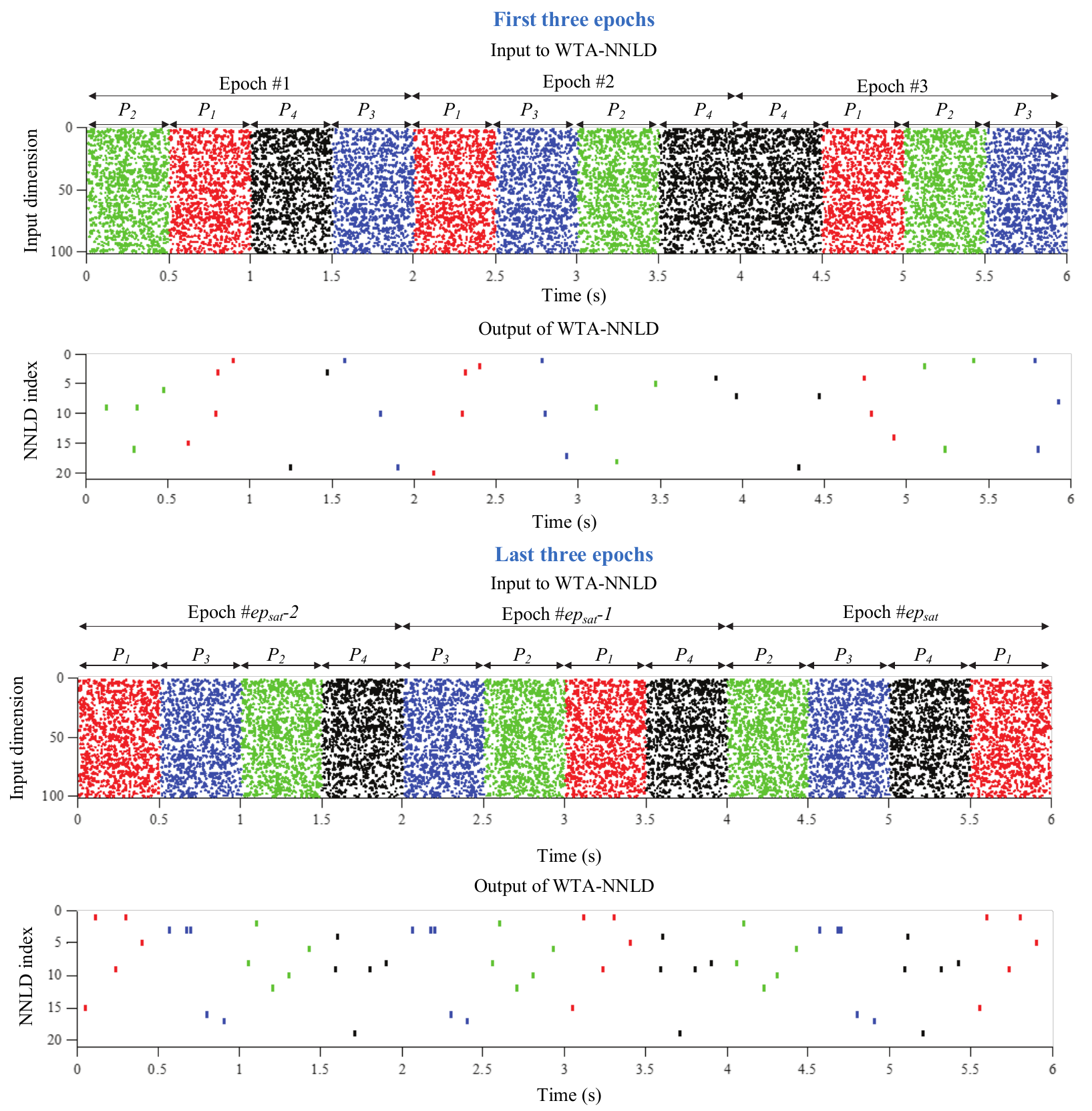}
		\caption{The input and output of WTA-NNLD has been shown for a particular trial of the four class classification when $n_{sub} = 5$. $P_1$, $P_2$, $P_3$ and $P_4$ represent the patterns of a particular class. The figure depicts that before learning WTA-NNLD produces arbitrary spikes whenever a pattern is presented. After learning, the network produces unique sequence of NNLD spikes for patterns of different classes. This unique sequence acts as an identifier of the pattern class.}
		\label{first3_last3_epochs}
	\end{center}
\end{figure*}

\subsection{Case 2: $n_{sub}>1$}
Next, we consider $n_{sub} = 5$ i.e. we divide each pattern into 5 subpatterns by setting $\tau_{s,inh}$ and $I_{0,inh}$ as per Equation \ref{tau_inh_I0_inh}. For $C$ class classification, the maximum number of subpatterns can be $C \times n_{sub} $ so we set $N = C \times n_{sub}$ in this case i.e. we keep $N$ = 10, 20 and 30 for two, four and six class classification respectively. Considering $\sigma_{jitter} = 0$, the evolution of $CM$ with epochs for two, four and six class classification averaged over 50 trials is shown in Fig. \ref{nsub_more_1}(b). Moreover, the value of $ep_{sat,avg}$ (averaged over 50 trials) is found out to be 210, 221 and 230 when $C$ = 2, 4 and 6 respectively. Unlike Case 1, here we consider the response to a pattern as a unique firing sequence of few NNLDs. As an example, we consider a particular trial of four class classification and look into the first and last 3 epochs during its training. It is evident from Fig. \ref{first3_last3_epochs} that during the first 3 epochs, WTA-NNLD produces arbitrary sequences of spikes. However, it can be seen that after the training of the network is complete, WTA-NNLD produces different firing sequences for different patterns while producing the same sequence when same patterns are encountered. WTA-NNLD trained by this method produces a 100\% accuracy in recognizing different patterns by producing its unique firing sequence for two, four and six class classification. The performance of the network with varying intensity of jitter is provided in Fig. \ref{n_sub_1}(c) (spikes present in all afferents) and Fig. \ref{nsub_more_1}(a) (spikes present in only half of the afferents) which depict that the $n_{sub}$ = 5 case is less jitter resilient than the $n_{sub}$ = 1 case.

We further increase the resolution of pattern subdivision by decreasing $\tau_{s,inh}$. We consider $n_{sub}$ = 10 and following the principle of $n_{sub}$ = 5, the number of NNLDs employed for $n_{sub} = 10$ are 20, 40 and 60 for two-class, four-class and six-class classification. This approach also provides 100\% accuracy in providing a unique sequence of firing whenever a particular pattern is encountered when $\sigma_{jitter} = 0$. However, the performance of the network falls rapidly with the increase of $\sigma_{jitter}$ as shown in Fig. \ref{n_sub_1}(c) and Fig. \ref{nsub_more_1}(a). We also show the evolution of $CM$ with the epochs in Fig. \ref{nsub_more_1}(b). Furthermore, the number of epochs needed for convergence of $CM$ in this case is much more than the previous cases as depicted in Fig. \ref{nsub_more_1}(c). We conclude that dividing a pattern into too many subpatterns hampers the network performance.

Next we delve a bit further and show the statistics of causes for the failure of the system in producing successful trials. A trial may fail if either condition (a) or (b) (described in Section \ref{spec_sens}) is not satisfied. We denote the failure of condition (a) as $F1$. Note that for a trial, condition (b) can fail if a pattern is misclassified as a pattern of another class (denoted as $F2$) or as a random pattern (denoted as $F3$).  Table \ref{fail_ana} shows the statistics of failed trials for $n_{sub}=$ 1, 5 and 10 when $\sigma_{jitter}/\tau_s \approx 0.1$. Note that $F1$ is high for $n_{sub}=$ 1 since a unique NNLD might lock onto multiple classes of patterns. $F1$ reduces for $n_{sub}=$ 5 and increases again for $n_{sub}=$ 10 since sometimes the network fails to produce a unique 10 indices long representation for all patterns of the same class.    

Moreover, we test our network with random patterns and note the cases where a learnt unique representation is produced for a random input pattern i.e. a false positive error occurs. The percentage of false positive errors produced for $n_{sub}=$ 1, 5 and 10 when $\sigma_{jitter}/\tau_s \approx 0.1$ are 8\%, 0\% and 0\% respectively. Note that no false positive errors occur for $n_{sub}=$5 and 10 since it is highly unlikely for a random pattern to make a sequence of neuron firing same as any of the learnt representation.
	
\begin{table}[!ht]
	\caption{Analysis of failure statistics}
	\centering
	\begin{tabular}{|c|c|c|c|c|}
		\hline
		Case & $n_{sub}=1$  &  $n_{sub}=5$   & $n_{sub}=10$ \\ \hline
		$F1$    &  12\%      &   2\%    & 6\%   \\ \hline
		$F2$    &   2\%		 &   2\%   &  2\%  \\ \hline  
		$F3$    &   2\%      &   4\%    & 8\%	\\ \hline
	\end{tabular}
	\label{fail_ana}
\end{table}

\section{VLSI Implementation: Effect of statistical variation}
In this section we analyze the stability of our algorithm to hardware nonidealities by incorporating the statistical variations of the key subcircuits. The primary subcircuits needed to implement our architecture are synapse, dendritic squaring block, neuron and $c^n_{pj}$ calculator. While the variabilities of the synapse circuit are modeled by mismatch in the amplitude ($I_0$) and time constant ($\tau_s$) of the synaptic kernel function, the variabilities of the squaring block are captured by a multiplicative constant ($cb_{ni}$) \cite{roy_tbcas}. We do not consider the variation of inhibitory current kernel since it is global and only a single instance is present in the architecture. In our earlier work \cite{roy_tbcas,amitava_iscas_2015}, we proposed the circuits for implementing the synapse and squaring block of NNLD and performed Monte Carlo analysis to find their variabilities. We presented that the $\frac{\sigma}{\mu}$ of $I_0$, $\tau_s$ and $cb_{ni}$ for the worst case scenario are 13\%, 10.1\% and 18\% respectively. The mismatch of the LIF neuron circuit proposed in \cite{amitava_iscas_2015} was captured by variations in the firing threshold $V_{thr}$, the $\frac{\sigma}{\mu}$ of which was computed to be 12.5\%. Lastly, the nonidealities of the $c^n_{pj}$ calculator block, described in \cite{RoyISIC2014}, are modeled as a multiplicative constant ($cc_{ni}$). Monte Carlo analysis of the $c^n_{pj}$ calculator block revealed that its $\frac{\sigma}{\mu}$ for the worst case is 18\%. 

\begin{figure}[!htbp]
	\centering
	\subfloat[]{\includegraphics[width=0.48\textwidth]{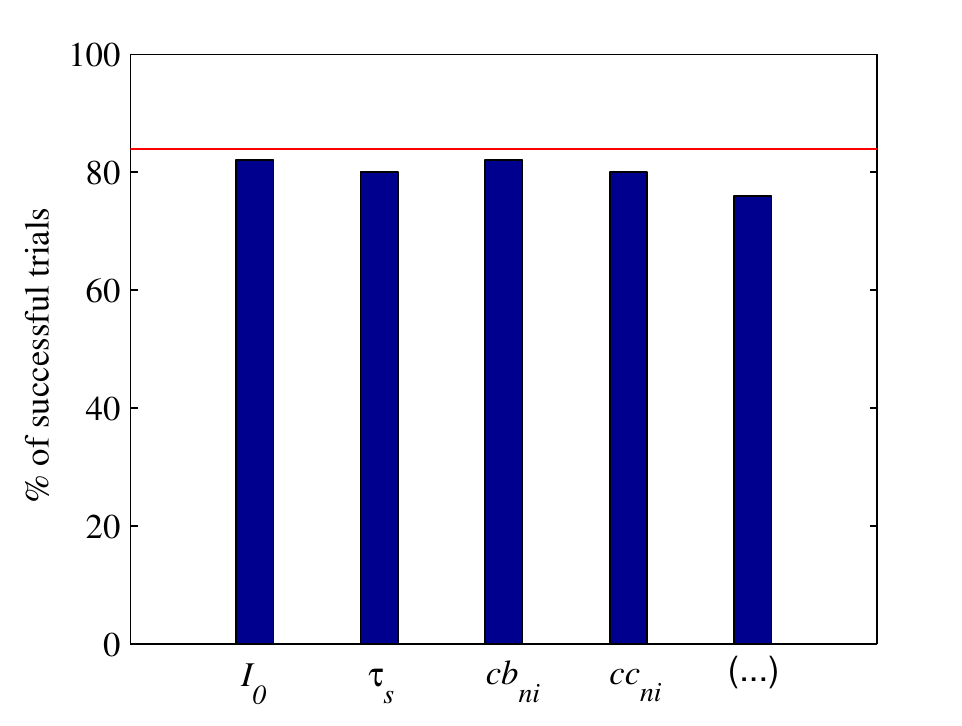}}\\
	\subfloat[]{\includegraphics[width=0.48\textwidth]{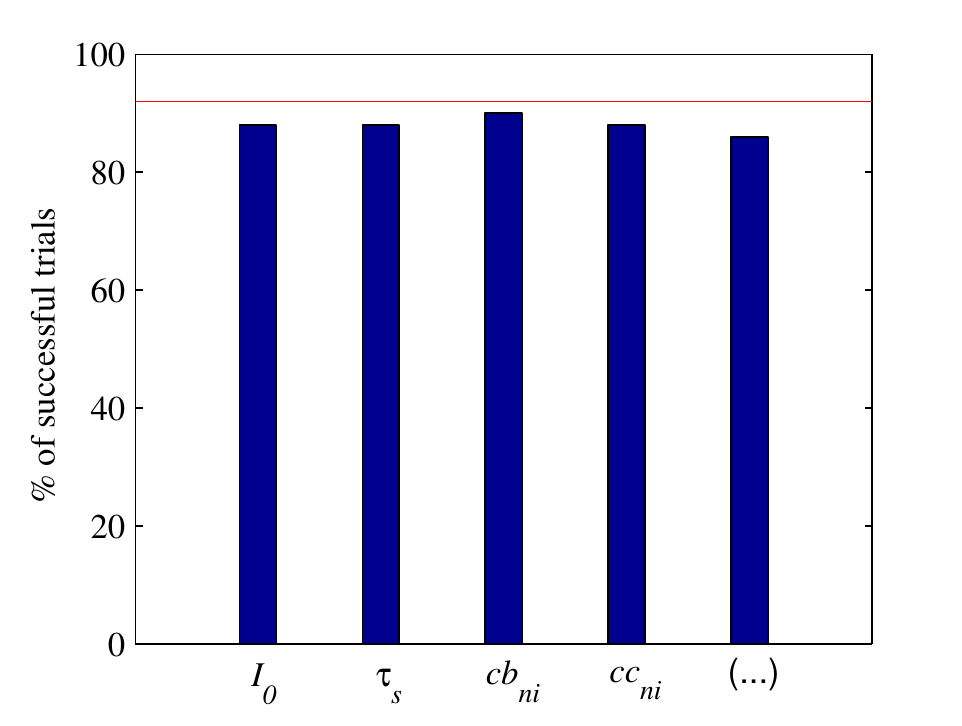}}\
	\caption{Stability of WTA-NNLD trained by STDP-NRW is plotted with respect to different hardware nonidealities for $n_{sub}=1$ (a) and $n_{sub}=5$ (b). The constant red line indicates the percentage of successful trials obtained by our method without any nonidealities when $\sigma_{jitter}/\tau_s \approx 0.1$. The bars represent the percentage of successful trials obtained after inclusion of nonidealities. The rightmost bar marked by (...) represents the performance when all the nonidealities are included simultaneously.}
	\label{unsup_nonid}
\end{figure}

Fig. \ref{unsup_nonid} shows the performance of the proposed method when these nonidealities are included in the model for $n_{sub}=1$ and $n_{sub}=5$ keeping $\sigma_{jitter}/\tau_s \approx $ 0.1. The bars corresponding to $I_0$, $\tau_s$, $cb_{ni}$, and $cc_{ni}$ denote the performance degradation when statistical variations of $I_0$, $\tau_s$, $cb_{ni}$, and $cc_{ni}$ are included individually. The results of Fig. \ref{unsup_nonid}(a) and Fig. \ref{unsup_nonid}(b) depict that the performance of the proposed algorithm is most affected by $\tau_s$ and  $cc_{ni}$ and least by $cb_{ni}$. Finally, to mimic the proper hardware scenario we consider the simultaneous implementations of all the nonidealities, which is marked by (...). The (...) bars show that there is an 8\% and 6\% decrease in performance for $n_{sub}=1$ and $n_{sub}=5$ respectively.

\section{Conclusion}
We have proposed a new neuro-inspired Winner-Take-All architecture (WTA-NNLD) and a STDP inspired dendrite specific structural plasticity based learning rule (STDP-NRW) for its training. Motivated by recent biological evidences and models suggesting nonlinear processing properties of neuronal dendrites we employ neurons with nonlinear dendrites to construct our WTA architecture. Moreover, we consider binary synapses instead of high resolution synaptic weights. Thus our learning rule, instead of weight updates, trains the network by modification of the connections between input and synapses. We have also provided a method by which the number of subpatterns per pattern learned by WTA-NNLD can be controlled. WTA-NNLD encodes patterns of different classes by either activity of distinct NNLDs or by a distinct sequence of NNLD firings. To demonstrate the performance of WTA-NNLD and STDP-NRW, we have considered two, four and six class classification of 100 dimensional Poisson spike trains. We can conclude from the result that the slow time constant of inhibitory signal ($\tau_{s,inh}$) can be properly set to obtain a tradeoff between specificity and sensitivity of the network. Our immediate future work will include studying the effects of connection changes after the network gets integrated over multiple patterns. This will reduce the number of required computations. On another note, we will look into the classification of spike based MNIST \cite{Lecun98,MNIST-DVS} datasets by our method. Our network can be immediately scaled to learn the digits of MNIST dataset, the only requirement being additional simulation time and computational memory compared to the tasks considered in this article. Furthermore, to achieve invariance to scaling and rotation during image classification, we will be constructing NNLD based convolutional neural networks \cite{LeCun1998CNN} trained by structural plasticity. We will also implement the proposed network in hardware and apply it for real time online unsupervised classification of spatio-temporal spike trains.


\bibliographystyle{IEEEbib}

\end{document}